\documentclass[a4paper,11pt,openany]{article}
\usepackage[utf8]{inputenc}
\usepackage{amsmath}
\usepackage{geometry}
\PassOptionsToPackage{showframe}{geometry}
\usepackage{graphicx}
\usepackage[english]{babel}
\usepackage{listings}
\usepackage{amssymb}
\usepackage[toc,page]{appendix}
\usepackage{supertabular}
\usepackage{longtable}
\usepackage{enumerate}
\usepackage[table]{xcolor}
\usepackage{array} 
\usepackage{float}
\usepackage{caption}
\usepackage{booktabs}
\usepackage{natbib}
\usepackage{multirow}
\setcitestyle{authoryear}
\usepackage{mathtools}

\usepackage{blindtext}
\usepackage{layaureo}

\graphicspath{{./Images/}}
\usepackage{url,hyperref}
\usepackage{color}
\usepackage{comment}
\usepackage{mathrsfs}  
\usepackage{amsfonts, amssymb}
\usepackage{algorithm,algorithmic}

\usepackage[normalem]{ulem}
\newcounter{ToDo}
\newcounter{gaocomm}
\newcounter{Note}
\definecolor{blue-violet}{rgb}{0.54, 0.17, 0.89}
\definecolor{mygreen}{rgb}{0.0, 0.5, 0.0}
\definecolor{awesome}{rgb}{1.0, 0.13, 0.32}
\definecolor{bostonuniversityred}{rgb}{0.8, 0.0, 0.0}

\usepackage{authblk}
\title{Differentiable Neural Architecture Search with Morphism-based Transformable Backbone Architectures}
\author[1]{Renlong Jie}
\author[1]{Junbin Gao}
\affil[1]{The University of Sydney\\
Camperdown NSW 2006}

\begin{document}
\maketitle

\begin{abstract}
 This study focuses on the highest-level adaption for deep neural networks, which is at structure or architecture level. To determine the optimal neural network structures, usually we need to search for the architectures with methods like random search, evolutionary algorithm and gradient based methods. However, the existing studies mainly apply pre-defined backbones with fixed sizes, which requires certain level of manual selection and in lack of flexibility in handling incremental datasets or online data streams. This study aims at making the architecture search process more adaptive for one-shot or online training. It is extended from the existing study on differentiable neural architecture search, and we made the backbone architecture transformable rather than fixed during the training process. As is known, differentiable neural architecture search (DARTS) requires a pre-defined over-parameterized backbone architecture, while its size is to be determined manually. Also, in DARTS backbone, Hadamard product of two elements is not introduced, which exists in both LSTM and GRU cells for recurrent nets. This study introduces a growing mechanism for differentiable neural architecture search based on network morphism. It enables growing of the cell structures from small size towards large size ones with one-shot training. Two modes can be applied in integrating the growing and original pruning process. We also implement a recently proposed two-input backbone architecture for recurrent neural networks. Initial experimental results indicate that our approach and the two-input backbone structure can be quite effective compared with other baseline architectures including LSTM, in a variety of learning tasks including multi-variate time series forecasting and language modeling. On the other hand, we find that dynamic network transformation is promising in improving the efficiency of differentiable architecture search.
 \end{abstract}
 
 \section{Introduction}\label{Sec:6.1}
Neural architecture search (NAS) enables the learning of model structures of artificial neural networks \citep{elsken2018neural}. In early years, neural network architectures are considered as discrete hyper-parameters, which can be encoded to sequences of discrete variables representing the combinations of neurons, cells and connections. The encoded vectors with corresponding model performances can then be handled and optimized by population based methods such as random search and evolution algorithm \citep{jaderberg2017population, real2017large, real2018regularized, stanley2002evolving, suganuma2017genetic}. However, traditional way of population-based methods can be computational expensive for NAS, which involves a large number of full model training for exploration. Therefore, it is necessary to make the most of exploitation with previous trials so that the searching process can be more efficient. Beyond evolutionary algorithms, methods like reinforcement learning \citep{zoph2016neural, pham2018efficient, zhong2018blockqnn}, Bayesian optimization \citep{pelikan1999boa} and network morphism \citep{wei2016network, cai2018efficient, cai2018path} can also be applied in reducing the searching cost. With the increased model size implemented in deep learning applications, exploring hundreds of architectures can still be a quite heavy load. In this background, few-shot of even one-shot NAS with gradient-based approaches has become a hot topic of study in recent years. 
\newline\newline
Gradient based NAS aiming at doing end-to-end architecture learning with gradient-based optimizers and back propagation. It can reduce the searching process into one-shot training with an over-parameterized architecture \citep{bender2018understanding, liu2018darts, cai2018proxylessnas}. One fundamental work is the invention of Differentiable Architecture Search scheme, in which over-parameterized graphs are initialized as the search space, while softmax functions are implemented for parameterizing the discrete architecture structures into differentiable ones \citep{liu2018darts}. However, as a deterministic method with all softmax weights kept during training, it requires a two-stage process for architecture training and validation. Stochastic methods can address this issue by replacing the weight vector for operation combination with one-hot or binary function, which can be effectively done by random sampling from Gumbel-Softmax distributions \citep{xie2018snas, chang2019data}. There are also methods based on Bayesian inference \citep{zhou2019bayesnas, shaw2019meta, zhou2020posterior}, natural gradient \citep{akimoto2019adaptive}, meta learning \citep{finn2017model, lian2019towards, elsken2020meta}, where the similar search space is implemented. Also there are works focusing on developing new search spaces \citep{mei2019atomnas}, which can contribute to an increasing of the efficiency of searching as well as model performance of the architecture being found. 
\newline\newline
There are also studies on introducing dynamic networks for architecture adaptation during model training, in which dynamical training and pruning can be made with gradient information \citep{dai2019nest, dai2019grow}. Moreover, further discussion on how to handle the splitting of gradient for growing dynamic neural networks has been raised \citep{wu2019splitting}, which provided a second-order method for learning lightweight neural architectures in resource-constrained settings. On the other hand, for network pruning, study has shown that knowledge transfer can be introduced with an improvement of training efficiency \citep{dong2019network}.
\newline\newline
The limitation of existing studies can be illustrated from three aspects: First, it is difficult to determine the size of over-parameterized architecture or the architecture search space for a particular task or dataset without exploratory or empirical study, which often requires expert knowledge and a large amount of extra computational cost, resulting in a reduction of the degree of automation. Second, there are only pruning mechanisms after setting the over-parameterized architecture, while it is possible that the initial architecture space is smaller than required, which means growing mechanism should be introduced for higher level of adaptation. Third, in most existing studies, each operation can only appear once for each node. However, for non-linear mappings, an ensemble of the same operation can provide an increase of expressiveness. 
\newline\newline
In this study, we propose a hybrid method of automatically growing the architecture during the learning process of gradient-based one-shot architecture search. We introduce the technique of network morphism to keep the learned mapping as much as possible in each growing transformation. The main contributions include:
\begin{itemize}
    \item We introduce both dynamic growing and pruning mechanisms into differentiable architecture search algorithms.
    \item We implement network morphorism for both node growing and dynamic operation transformation. 
    \item We implement a two-input backbone architecture for RNNs and corresponding two-input operation set.
    \item We introduce two pruning mechanisms for the process of cell structures growing.
\end{itemize}

\section{Related works}\label{Sec:6.2}
\subsection{Differentiable architecture search}\label{Sec:6.2.1}
Gradient based neural architecture search largely reduces the computational cost in finding architectures with high performance. By following the idea of existing differentiable neural architecture search literature \citep{liu2018darts, cai2018proxylessnas}, we represent the search space with a directed acyclic graph (DAG), in which the output of a mixed operation $o_{ij}$ can be calculated as follows:
\begin{equation}
    o_j(z_i) = \sum_{o\in O}w_{ij}^o o_{ij}(z_i)
    \label{Eq:6.1}
\end{equation}
where $i<j$ and $z_j$ can be obtained as $\sum_{i<j}o_j(z_i)$. To facilitate normalization and back propagation procedure, we can use parameterized Softmax functions to represent $w^o_{ij}$:
\begin{equation}
    w^o_{ij} = \frac{\exp(\alpha_o^{i,j})}{\sum_{o'\in O}\exp(\alpha_{o'}^{(i,j)})}
    \label{Eq:6.2}
\end{equation}
where $\alpha_o^{i,j}$s are the parameters that can be trained with back-propagation. In the original study of DARTS \citep{liu2018darts}, the available operations include linear transformations followed by one of tanh, relu, sigmoid activations, as well as identity mapping and zero operation. With the over-parameterized architecture along with the trainable architecture parameters, the bi-level optimization problem can be given by Eq.~\eqref{Eq:6.3}.
\begin{equation}
    \begin{split}
        &\min_{\alpha}L_{\text{val}}(w^{*}(\alpha), \alpha)\\
        &\text{s.t. }w^{*}(\alpha) = \text{argmin}_{w} L_{\text{train}}(w, \alpha)
    \end{split}\label{Eq:6.3}
\end{equation}
The second sub-optimization is generally a normal training process of model parameters with training data, while the first one is optimized on validation set and can be done with finite difference approximation. The full algorithm of DARTS is given by Algorithm \ref{Alg:4.1}.
\begin{algorithm}[!htb]
\caption{Algorithm for DARTS}
\begin{algorithmic}[1]
  \STATE Create a mixed operation $\bar{o}_{i,j}$ parameterized by $\alpha^{(i,j)}$ for each edge $(i,j)$.
  \WHILE{not converged}
  \STATE Update architecture $\alpha$ by descending $\nabla_{\alpha}L_{\text{val}}(w-\xi\nabla_{w}L_{\text{train}}(w,\alpha),\alpha)$ ($\xi=0$ if using first order approximation)
  \STATE Update weights $w$ by descending $\nabla_w L_{\text{train}}(w,\alpha)$.
  \ENDWHILE
  \STATE Derive the final architecture based on the learned $\alpha$.
\end{algorithmic}
\label{Alg:4.1}
\end{algorithm}

For architecture parameter updating, we need to implement gradient of validation loss with respect to architecture parameter $\alpha$. The second order approximation is given by:
\begin{equation}
\begin{split}
    \nabla_{\alpha}L_{\text{val}}(w^{*}(\alpha), \alpha)
    \approx \nabla_{\alpha}L_{\text{val}}(w-\xi\nabla_{w}L_{\text{train}}(w, \alpha), \alpha)
    \end{split}
\end{equation}
Applying Tylor's expansion, it can be further approximated by:
\begin{equation}
    \nabla_{\alpha}L_{\text{val}}(w^{'}, \alpha)-\xi \nabla^2_{\alpha,w}L_{\text{train}}(w, \alpha)
    \nabla_{w^{'}}L_{\text{val}}(w^{'}, \alpha)
\end{equation}
where $w'=w-\xi \nabla_{w}L_{\text{train}}(w, \alpha)$. The computational complexity of the second term can be reduced by finite difference approximation:
\begin{equation}
    \nabla^2_{\alpha,w}L_{\text{train}}(w, \alpha)
    \nabla_{w^{'}}L_{\text{val}}(w^{'}, \alpha) \approx \frac{\nabla_{\alpha}L_{\text{train}}(w^{+}, \alpha) - \nabla_{\alpha} L_{\text{train}}(w^{-}, \alpha)}{2\epsilon}
\end{equation}
where $w^{\pm} = w\pm\epsilon \nabla_{w^{'}}L_{\text{val}}(w^{'}, \alpha)$. Existing study has shown that first-order approximation leads to empirically worse performance \citep{liu2018darts}.
\subsection{Network morphism}\label{Sec:6.2.2}
Network morphism is defined as follows: After morphing a parent network, the child network is expected to inherit the knowledge from its parent network and also has the potential to continue growing into a more powerful one with much shortened training time \citep{wei2016network}.
In this study, we will introduce the idea of network morphism to the growing process of backbone architecture. Consider layer-structure \texttt{net2deeper} network morphism in non-linear case \citep{chen2015net2net}. Let $\phi$ be a non-linear activation function. Now assume we have two layers $B_l$ and $B_{l+1}$ and want to insert a new layer but keep the mapping function. That is, 
\begin{equation}
    B_{l+1} = \phi(G \circ B_l) = \phi( G\circ\phi^{'}(G^{'} \circ B_l))
\end{equation}
where $G$ and $G^{'}$ are usually affine transformations by weight matrices, and $\phi^{'}$ is the activation function in the newly inserted layer. Usually we need to reinitialize $G^{'}$ and $G$ to achieve the same mapping, while ensuring the effective of further training with a larger model. In Section ~\ref{Sec:3.2.1}, we will define the network morphism for backbone architectures in adding new nodes.

\subsection{Steepest descent splitting}\label{Sec:6.2.3}

Although network morphism can be applied to preserve the mappings after growing the architecture, it does not make the most of gradient information of previous training. Steepest descent splitting addresses this issue by reinitialize the parameter after growing with steepest descent technique \citep{wu2019splitting}, which aims at finding the updating that minimize the loss rather than simply taking the first-order gradient of loss. Assume the loss has a general form:
\begin{equation}
    L(\theta):=E_{x\sim D}[\Phi(\sigma(\theta,x))]
\end{equation}
where $\theta$ is the related model parameter of a neuron $\sigma$ and $x$ is the input variable. $D$ represents the data distribution, and $\Phi$ is a map determined by the overall loss function. If we split the neuron into $m$ off-springs, we have the augmented loss function:
\begin{equation}
    L(\theta, w):=E_{x\sim D}[\Phi(\sum_{i=1}^m w_i\sigma(\theta_i,x))]
\end{equation}
where $\theta:={\theta_i}^m_i$ are the parameters for different off-springs after splitting, and $\{w_i\}^m_{i=1}$ are the combination weights for the outputs of off-spring neurons. We define the splitting matrix:
\begin{equation}
        S(\theta) = E_{x\sim D}[\Phi'(\sigma(\theta,x))\nabla^2_{\theta\theta}\sigma(\theta,x)]
        \label{Eq:split}
\end{equation}
which is different from the Hessian matrix with the following relationship: 
\[\nabla^2_{\theta\theta}L(\theta)=S(\theta)+E[\Phi''(\sigma(\theta,x))(\nabla_{\theta}\sigma(\theta,x))^2]\] 
We call the minimum eigenvalue $\lambda_{\text{min}}(S(\theta))$ of $S(\theta)$ the splitting index of $\theta$, and the eigenvector $v_{\text{min}}(S(\theta))$ related to $\lambda_{\text{min}}(S(\theta))$ the splitting gradient of $\theta$. Then the steepest descent method could provide an improvement with respect to standard parametric update without splitting when $\lambda_{\text{min}}(S(\theta))<0$. Consider the case that we split a neuron into two off-springs, the optimal splitting strategy subject to $||\delta_i||\leq 1$ is given by
\begin{equation}
\begin{split}
&m=2\quad w_1=w_2=1/2\\
&\theta_1 = \theta-v_{\text{min}}(S(\theta))\quad \theta_2 = \theta+v_{\text{min}}(S(\theta))
\end{split}
\end{equation}
This result is guaranteed by Theorem 2.3 in \citep{wu2019splitting}. In this study, we can use steepest descent split for growing of operation in each node. 

\section{Method}\label{Sec:6.3}
In this study, we propose an approach for gradient-based one-shot neural architecture search by continuously growing the cell structure based on network morphism. The main idea can be summarized by the following steps:
\begin{itemize}
\item Construct a backbone with certain prior constraints. In this study we implement a recently proposed backbone where each node has two inputs.
\item Monitor the validation loss on learning tasks.
\item Increase the cell size with one node as the the training of the previous cell no longer reduces the validation loss. 
\item Inherit the trained weight of the previous cell architecture and introduce network morphism to keep the mapping unchanged after adding the new node.
\item Stop the whole process when the model after adding a new node does obtain an increase in terms of validation or test model performance. 
\end{itemize}
\subsection{Extendable backbone architecture} \label{Sec:6.3.1}
\begin{figure}[th] 
\begin{center}
 \includegraphics[width=1.0\linewidth]{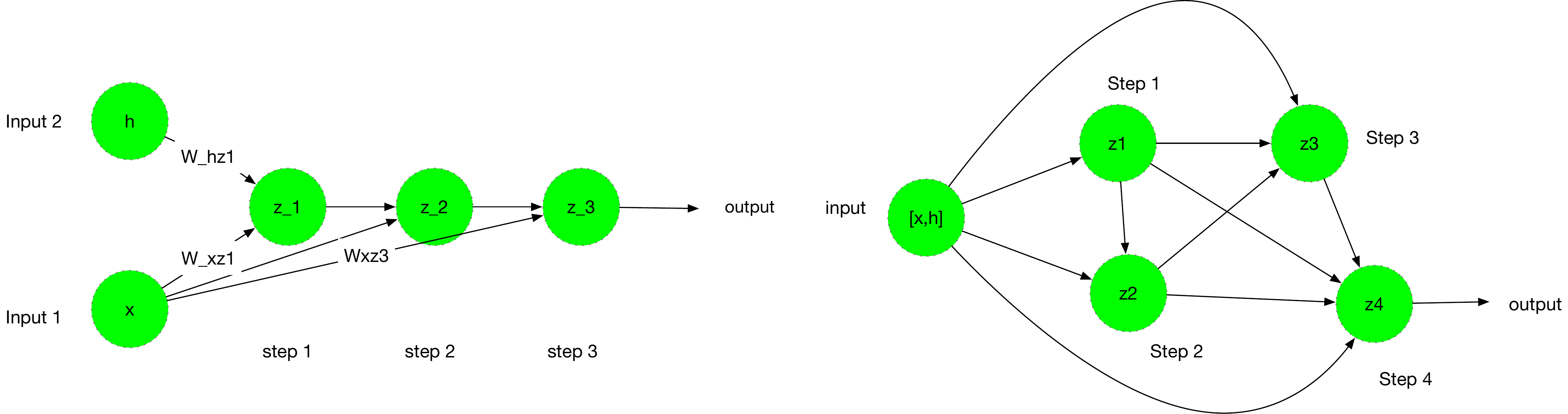}
 \caption{The diagram of the two-input backbone architecture and DARTS backbone with architecture growing.} \label{Fig:5.1}
\end{center} 
\end{figure}

First, we implement a novel backbone architecture with two inputs and one output, which is motivated by the cell structures used in RNN models and originally proposed in \citep{jie2021differentiable}. The diagram of the backbone architecture can be shown in Figure~\ref{Fig:5.1}.
Different from the backbone structure in DARTS, \citep{jie2021differentiable} introduce a new one in which each node can only have two input nodes, and each operation takes both of them. Therefore, we can notate it as \textbf{Two-to-One backbone}. The corresponding formula for information flow in each cell is given by Eq.~\eqref{Eq:6.4a}, \newline
\begin{equation}
\begin{split}
    o_j(x_t, h_{t-1}) &= \sum_{o\in O}w^o_{ij}o_{ij}(x_t, o_{j-1}(x_t, h_{t-1})),\, j=0,1,2,...\\
    ...\\
    o_0(x_t, h_{t-1}) &= \sum_{o\in O}w^o_{i0}o_{i0}(x_t, h_{j-1}))
    \end{split}\label{Eq:6.4a}
\end{equation}
where $w^o_{ij}$'s are given by softmax functions a trainable vector as is in Eq.~\eqref{Eq:6.2}. One advantage of the proposed backbone is that it provides a relatively straightforward way for introducing Hadamard product and other types of interaction operations that cannot be handled by concatenating all inputs as a single matrix. In addition, it can also tell apart the effects of hidden state from the last step with the input of the current step, and keep skip connections for both of them. For recurrent neural networks, where each cell has two inputs and one output, it is straightforward to build hierarchical structures based on this backbone architecture \citep{liu2018hierarchical}. However, as we add constraints on the number of inputting paths to each node, the nodes in the backbone architecture is not fully connected with all previous nodes. For DARTS, the operations implemented for RNN models include sigmoid, tanh, relu activations, identity mapping and zero operation. On contrast, for the proposed two-input backbone, a corresponding set of two-inputs operations are listed in Table~\ref{tab:5.1}.
\begin{table}[!htb] 
  \centering
  \caption{Operations in Two-to-One backbone architecture}
    \begin{tabular}{cc}
    \toprule
    Operation & Formula \\
    \midrule
    sigmoid & $o_{\sigma}(x_t, h_{t-1}) = \text{sigmoid}(W^s_{xh}x_t+W^s_{hh}h_{t-1}+b_s)$ \\
    tanh  &  $o_{\tanh}(x_t, h_{t-1}) = \tanh(W^t_{xh}x_t+W^t_{hh}h_{t-1}+b_t)$ \\
    relu  &  $o_{\text{relu}}(x_t, h_{t-1}) = \text{relu}(W^r_{xh}x_t+W^r_{hh}h_{t-1}+b_r)$ \\
    sum   &  $o_{\text{sum}}(x_t, h_{t-1}) = W^{su}_{xh}x_t + h_{t-1}$ \\
    hadamard product & $o_{\text{prod}}(x_t, h_{t-1}) = W^{p}_{xh}x_t*h_{t-1}$ \\
    \bottomrule
    \end{tabular}%
  \label{tab:5.1}%
\end{table}%

In the table, we assume $x_t$ and $h_{t-1}$ as the inputs. As we can see, during the learning and growing process, each newly added node can only be connected with the latest added node and the input. This designing also borrows idea of identity mapping from ResNet \citep{he2016deep}. We introduce skip connections between the output node and input, while for the hidden states, the summation operation given in Table~\ref{tab:5.1} can keep their identity mappings. To make up the defect of the reduced backbone architectures, we introduce online network transformation allowing the changing of backbone architecture as well as the operation list during the training process.

\subsection{Network morphism for architecture transformation} \label{Sec:6.3.2}
Beyond the existing studies on the growing of the number of layers and number of hidden units with net-to-wider or net-to-deeper transformations, we further introduce the growing of cell structure and the growing of operator with network morphism. On the other hand, we also introduce the transformation of operational replacing and re-sampling with replacement.  

\subsubsection{Growing of cell structure} \label{Sec:3.2.1}
For network morphism, after a new node is added, we need to make sure that the output of the new node in the re-initialized model is close to the output of the previous node for any input data. For the Two-to-One backbone architecture, the corresponding relationship is given in Eq.~\eqref{Eq:6.5}.
\begin{equation}
\begin{split}
    o_{j+1, \text{init}}(x_t, o_j) &= \sum_{o\in O}w^o_{ij}o_{ij}(x_t, o_{j}(x_t, o_{j-1})) \\
    &\approx o_{j}(x_t, o_{j-1})
    \end{split}\label{Eq:6.5}
\end{equation}
where $x_t$ is the input of the cell structure at position $t$ of the input sequence, $w^o_{ij}$ is the weight of $i$th operation in node $j$, which does not depend on the position $t$. Also, we share the notation of the output of each node $o_j$ and corresponding function of the two inputs. To achieve the approximation relationship in Eq.~\eqref{Eq:6.5}, we initialize the connection weights between the last node and the newly added node close to an identity mapping, while initializing the connection weights between the input and the new node close to zero. For example, if we have a set of operations given in Section~\ref{Sec:6.3.1}, then the combination can be written as:
\begin{equation}
\begin{split}
o_j(x_t, o_{j-1}) &= w_1 o_{\sigma}(x_t, o_{j-1}) + w_2 o_{\tanh}(x_t, o_{j-1}) + w_3 o_{\text{relu}}(x_t, o_{j-1})\\
&+ w_4 o_{\text{sum}}(x_t, o_{j-1}) + w_5 o_{\text{prod}}(x_t, o_{j-1})
\end{split}
\end{equation}
where we do not have identity mapping. In this case, the equality of Eq.~\eqref{Eq:6.5} holds when $w_1 = w_2 = w_3= w_5=0$ and $w_4=1$, while the corresponding $W^{su}_{xh}$ is a zero matrix. That is, we only keep summation operation with non-zero weight, in which the weights for the input $x$ is close to a zero matrix, so that the output of the last node will approximately to be the output of the newly added node. However, to make the updated model trainable after growing with back propagation, we need to add a random noise to zero weights to avoid zero gradients. Assume we have the $j$-th node in a trained backbone architecture, and then the $(j+1)$-th node can be added with the following re-initialization:
\begin{equation}
\begin{split}
    w_{j+1, k} & = \epsilon_{j+1, k},\, \epsilon_{j+1, k} \sim U(0,\delta),\, k=1,2,3,5\\
    w_{j+1, 4} &= 1-\epsilon_{j+1, 4}, \, \epsilon_{j+1, 4} = \sum_{i\in{1,2,3,5}}\epsilon_{j+1, k}\\
    W^{su}_{j+1, xh} &= (\epsilon_{pq})_{1\leq p\leq n_x, 1\leq q\leq n_h},\,\epsilon_{pq} \sim N(0,\sigma^2)\text{ for }\forall{p,q}
    \end{split}
\end{equation}
where $w_{j+1, 4}$ is the combination weight for the sum operation in the newly added node, $w_{j+1, k}$ represents the weights for other operations, while the corresponding parameter $\alpha_{j+1, k}$ can be obtained by inverting the Softmax functions based on the re-initialized weights:
\begin{equation}
    \alpha_{j+1, k} = \ln(w_{j+1, k})+\ln(\sum_k \exp(\alpha_{j+1, k})).
\end{equation}
Still, $W^{su}_{j+1, xh}$ is the weight matrix connecting the input and the new output in the summation operation, which could be a Gaussian random matrix $(\epsilon_{pq})_{1\leq p\leq n_x, 1\leq q\leq n_h}$. The noise level is controlled by $\sigma$.

On the other hand, for DARTS backbone, we can also introduce growing of nodes with network morphism. Consider $z_j = \sum_{i<j}o_j(z_i)$ with $o_j(z_i) = \sum_{o\in O}w_{ij}^o o_{ij}(z_i)$, when a new node is added to the backbone, to keep the same mapping, we need:
\begin{equation}
    z_{j+1} = \sum_{i<j+1}o_{j+1}(z_i) = z_j
\end{equation}
Notice that $j$ is the index of the last node, while the summation is across all the previous nodes. This requires zeros operation for the connection between the newly added operation and each node except for the last node. Therefore, one solution can be demonstrated as this: Assume that the indices for the identity mapping and zero mapping are 4 and 5 in DARTS backbone operator. For previous nodes, we set the weight for zero operation close to 1, and the weights for other operations close to 0 with certain level of randomness: 
\begin{equation}
\begin{split}
    w_{j+1, k} & = \epsilon_{j+1, k},\, \epsilon_{j+1, k} \sim U(0,\delta),\, k=1,2,3,4\\
    w_{j+1, 5} &= 1-\epsilon_{j+1, 5}, \, \epsilon_{j+1, 5} = \sum_{i\in{1,2,3,4}}\epsilon_{i+1, k}
    \end{split}
\end{equation}
For the last node, set the weight for identity operation close to 1, and the weights for other operations close to 0 with certain level of randomness:
\begin{equation}
\begin{split}
    w_{j+1, k} & = \epsilon_{j+1, k},\, \epsilon_{j+1, k} \sim U(0,\delta),\, k=1,2,3,5\\
    w_{j+1, 4} &= 1-\epsilon_{j+1, 4}, \, \epsilon_{j+1, 4} = \sum_{i\in{1,2,3,5}}\epsilon_{i+1, k}
    \end{split}
\end{equation}
There are one or two extra hyper-parameters introduced by these transformation, which are $\delta$ for DARTS backbone and $\delta$ and $\sigma$ for Two-to-One backbone. In practice, since both $\delta$ and $\sigma$ are implemented as distribution parameters for sampling small values, it is not harmful to simply set $\delta = \sigma$ in most cases.


\subsubsection{Growing of operators} \label{Sec:3.2.2}
Under the the Two-to-One backbone architecture, the growing of cell structure will actually increase the distance between the input of step $t-1$ and the output of step $t$. We can consider this as vertical growing or the growing of nodes. On the other hand, we also propose a horizontal growing in terms of operators, which means we can add more component operations into each operator at each node of the cell structure. Comparing with the existing study of net to wider net or net to deeper net \citep{chen2015net2net}, the newly proposed growing technique is within cell level and can be more flexible in the context of recurrent neural networks. This can also be introduced to DATRS backbone. It provides the solution of both using different number of nodes and applying different numbers of operations in different nodes, while solely increasing the number of hidden states or the number of stacked RNN layers can be much less expressive. Although there is existing work also considering implementing arbitrary number of operations in each node \citep{chang2019data}, there is no mechanism of having multiple non-linear operations within one node.
\newline\newline
In the proposed method of operators growing, the newly added operation is the operation with the highest weights or probability being sampled, following particular criterion. That means we may have the combination of two or more same operations in each node. Consider the case of two sigmoid functions: 
\begin{equation}
\begin{split}
o_{\sigma_a}(x_t, h_{t-1}) &= \sigma(W^a_{xh} x_t + W^a_{hh}h_{t-1}+b^a_h)\\
o_{\sigma_b}(x_t, h_{t-1}) &= \sigma(W^b_{xh} x_t + W^b_{hh}h_{t-1}+b^b_h)
\end{split}
\end{equation}
where we take a single-node cell for example with $o_j=o_0=h_{t-1}$. Since the weighted combination of these two operations can not be fully expressed by one operation with certain weights in any circumstances, this will actually increase the model expressiveness. It can be considered as making an ensemble of layers with shared output activation. 
\newline\newline
For the growing of operators in general, we have two strategies: The first is to achieve network morphism approximately by initializing small combination weights and adjust the weights of other existing operations. On the other hand, we can initialize random parameters within the newly added operation. The second strategy is to apply steepest descent splitting technique, which is developed in \citep{wu2019splitting}. Since usually we use same number of hidden state in each gates for RNN cells, the splitting of operations require splitting all the hidden unit in this operation simultaneously. Assume that the loss before and after split are:
\begin{equation}
    \begin{split}
    L_0(\theta) &= L(\Phi(\sum_{i=1}^m \omega_i o_i(x, \theta_i))\\
        L_1(\theta) &= L(\Phi(\sum_{i=1, i\neq l}^m \omega_i o_i(x, \theta_i)+ (\omega_{l1}o_l(x, \theta_{l1})\\
        &+\omega_{l2}o_l(x, \theta_{l2}))))
    \end{split}
\end{equation}
where $m$ is the number of operations in the current node, $l$ is the index of operation to be split, $x$ is the input to the node, and $\theta_i$ is the possible parameters in for operation $i$. Since each operation in each node has multiple hidden states, each particular $\theta_i$ is usually a 2D matrix with size $h\times h$, $(h+x)\times h$ or $(h+x)\times (h+x)$ rather than a vector. To split these weight matrices, we can treat each column as a parameter vector of one hidden state, and then the splitting matrix of this hidden state can be calculated by Eq.~\eqref{Eq:split}. Finally we will have one extra operation with the same number of hidden states, while the parameters for each hidden unit is split from its counterpart of the original unit. 
\newline\newline
Along with the growing of nodes, the growing of operators can increase the model capacity during the learning process even without changing the number of hidden state and the number of stacked layers. If we allow multiple operations to be kept in each path towards each specific node in the final model, it is possible to initialize a relatively small number of hidden states, and make the cell structure growing the capacity automatically with fixed number of nodes.

\subsubsection{Pruning} 
Pruning technique is widely implemented in existing studies of neural architecture search, especially in the original study of DARTS \citep{liu2018darts}, where each node will only retain the top-k strongest operations ($k=1$ for RNN and $k=2$ for CNN). When growing backbone architectures are implemented, it is possible to do model pruning or operation selection after each stage of training before adding new nodes. Corresponding mechanism will be introduced in Section ~\ref{Sec:6.3.3}. Moreover, in this study we also introduce the dynamic pruning process based on the combination weights of operations, which means when the weight for a candidate operation is less than a certain threshold, the corresponding operation will be removed and weights will be reallocated for other operations. With dynamic growing and pruning mechanisms of operations, architecture transformation can be more adaptive to the training process as well as the learning tasks by making use of weights information. 


\subsubsection{Replacing and Re-sampling}
In this study we further propose a replacing technique, where we can keep a certain number of candidate operations in each nodes, but continuous replacing the operations with smaller weights to that with large weights. Since we can train the combination differently for different nodes, this will bring more flexibility and enlarge the architecture space that can be learned. Also we can release the number of operations in each node and determine which combination will be used by re-sampling. 

\subsubsection{Criterion for architecture transformation}
The parameters of fully trained models with smaller architectures will be applied for the updated model with new node. There are existing studies on training a meta-controller as a Reinforcement learning agent to determine the action for network transformation \citep{cai2018efficient}. However, as we focus on one-shot dynamic architecture search, there are no sufficient trial samples for training the agent. Instead we propose a set of criterion with as few as possible extra hyper-parameters for backbone structure evolving. One idea is to implement early stopping or convergence criterion to determine the time point and inheritable parameters from the trained models. In addition, we can apply thresholds on the combination weights for determine when to update the combination of operations in each operator. The criterion implemented in this study including the following items: 
\begin{itemize}
    \item Criterion for growing of nodes: We implement two-level early stopping criteria to determine when a new node will be added in the architecture, and when the whole growing process should be stopped. That is when the validation performance during the training of backbone architecture is no longer increasing for several epochs, we add a new node to the backbone. When adding new node no longer increases the model performance on validation set, when stop the whole growing process and select the optimal architecture. Similar criterion for architecture updating is also implemented in \citep{wu2019splitting}.
    \item Criterion for operator growing: We add an other operation that has the largest weights when its weights surpass a certain threshold under the strategy of network morphism. Under steepest descent strategy, when the validation loss is no-longer decreasing, we can select the operation with the largest number of negative minimum eigen values of splitting matrices in their parameter vectors. If all the hidden states have negative minimum eigen values for multiple operations in each path, we can split these operations simultaneously or split the one with minimum summation of minimum eigen values. These can be considered as different settings.
    \item Criterion for operator pruning: We prune an operation in a node when its corresponding weight is close to 0, while the weight of the operation with the largest weight does not surpass a threshold. 
\end{itemize}

\subsection{Two modes for architecture growing procedure}\label{Sec:6.3.3}
Without implementing dynamic transformation on operations, existing works do architecture pruning after the training of architecture parameters. In the setting with the growing of nodes, when the current model converges, it is possible to prune the architecture first before adding a new node. This results in two strategies to handle the architectures pruning during or after searching.

\textbf{Two strategies}:
\begin{itemize}
\item The first one is ``pruning at each stage'', which means we prune the architecture by only keeping the selected operators in after each stage of training. 
\item The second one is ``pruning at the end'', which means we use the expanded architecture after each stage for growing, while pruning will only happen when we finish the whole growing process.
\end{itemize}
\begin{figure}[th] 
\begin{center}
 \includegraphics[width=1.0\linewidth]{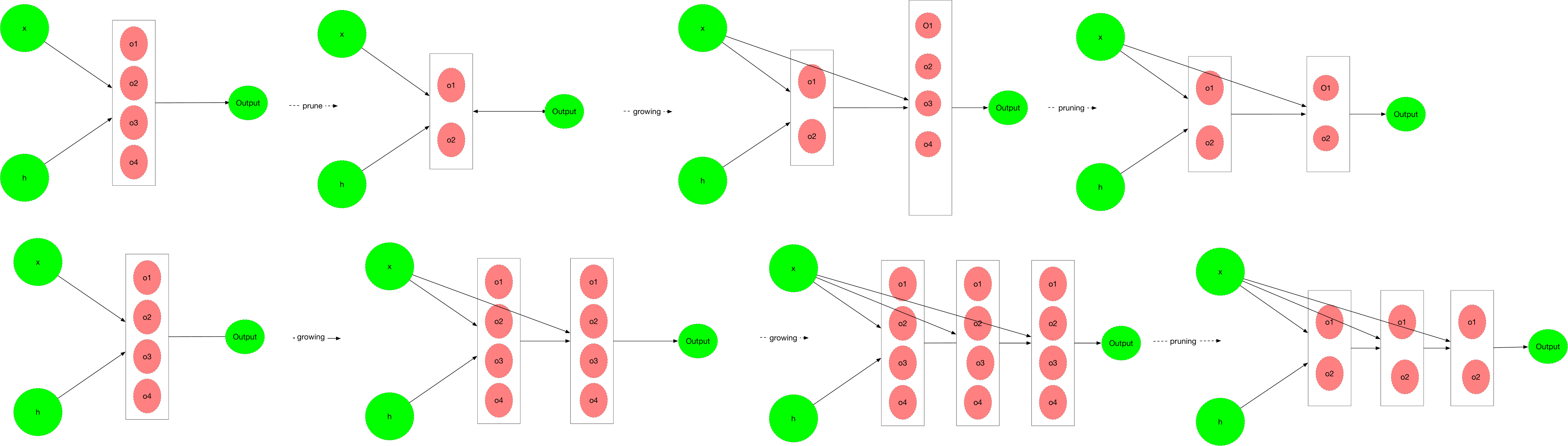}
 \caption{The diagram of two mode of architecture growing} 
\end{center} \label{Fig:5.2}
\end{figure}
Notice that ``pruning at the end'' is the one implemented in the original study of DARTS, where there is no growing mechanism for cell structure. Here we introduce ``pruning at each stage'' as another strategy when growing mechanism is implemented. In practice, if we assume an optimal number of operations in each node, the pruning process will base on that. On the other hand, the proposed dynamic operation transformation based on combination weights does not have a constraints on the number of operations, where the corresponding hyper-parameter is the threshold for growing or pruning.  Therefore, we can consider dynamic operation transformation and these two pruning strategies as three different training mechanisms. 

\subsection{Full algorithm}\label{Sec:6.3.4}
The algorithms introduced in this study are composed of three elements. The first is the general algorithm of the whole searching process, the second is the updating mechanism for node growing, and the third is the updating mechanism for dynamic operation transformation, which are demonstrated as pseudo codes in Algorithm~\ref{Alg:4.2}, Algorithm~\ref{Alg:4.3} and Algorithm~\ref{Alg:4.4}. In Algorithm~\ref{Alg:4.2}, \texttt{train(A,D)} is the training function that takes the initialized architecture $A$, training data $D_t$ and validation data $D_v$, which corresponds to the backbone training in DARTS. \texttt{Prune(A,$\theta$)} is the pruning function that receives the original architecture as well as the trained model parameters, and returns the pruned architecture. \texttt{Tune(A,$D_t$)} is the tuning function that receives the pruned architecture $A$, training data $D_t$ model parameters $\theta$, and returns the fine-tuned model parameters of pruned architecture after convergence. \texttt{Evaluate(A,$D_t$)} is the function that generate the evaluation performance on test set $D_{\text{test}}$ with trained architecture and model parameters. The most essential component is the updating function \texttt{update(A,$\theta$)}, which is implemented for growing the architecture and model parameters $(A, \theta)$ to new ones with techniques like network morphism and steepest descent. The details for the updating function are given by Algorithm~\ref{Alg:4.3} and Algorithm~\ref{Alg:4.4}. 
\begin{algorithm}
\caption{The Algorithm of Growing Architecture Search}
\begin{algorithmic}[1]
  \REQUIRE Initial architecture $A_0$, $\theta_0$
  \STATE Set $\text{eva\_best}=\text{a large number}$
  \STATE $A = A_0, \theta = \theta_0$
\FOR{$i$ in $1,2,3,...,S$}
  \STATE Backbone training: $A, \theta = \text{train}(A, D_t, D_v)$
  \IF{Pruning at each stage}
  \STATE Pruning: $A = \text{Prune}(A, \theta)$
  \STATE Fine tuning: $\theta = \text{Tune}(A, D_t, \theta)$
  \ENDIF
  \STATE Evaluate: $\text{eva} = \text{Evaluate}(A, \theta, D_{test})$
  \IF{$\text{eva}<\text{eva\_best}$}
  \STATE $A_{\text{best}}=A$
  \STATE $A, \theta=\text{update}(A, \theta)$
  \STATE $\text{eva\_best} =\text{eva}$
  \ELSE
  \STATE break
  \ENDIF
  \ENDFOR
\IF{Prune at the end}
  \STATE Pruning: $A = \text{Prune}(A_{\text{best}}, \theta)$
  \STATE Evaluate: $\text{eva} = \text{Evaluate}(A, D_{test})$
 \ENDIF
\end{algorithmic}
\label{Alg:4.2}
\end{algorithm}
\begin{algorithm}
\caption{The Algorithm of update function of nodes}
\begin{algorithmic}[1]
  \REQUIRE Previous architecture $A_{\text{prev}}$.
  \REQUIRE Previous model parameters $\theta_{\text{prev}}$.
  \STATE Initialize expanded architecture $A_{\text{expanded}}$.
  \STATE Inherit the parameter from the previous model for existing nodes.
  \STATE Initialize the newly added parameters in the expanded model with network morphism discussed in Section ~\ref{Sec:3.2.1}.
  \RETURN $A_{\text{expanded}}$, $\theta_{\text{expanded}}$
\end{algorithmic}
\label{Alg:4.3}
\end{algorithm}

\begin{algorithm}
\caption{The Algorithm of update function of operations}
\begin{algorithmic}[1]
  \REQUIRE Previous architecture $A_{\text{prev}}$
  \REQUIRE Previous model parameters $p_{\text{prev}}$
  \IF{Threshold of evolving operations reached during the training process}
  \STATE Select the operation to be duplicated by checking absolute combination weights or the number/summation of negative minimum eigen values for splitting matrices.
  \STATE Initialize expanded operations.
  \STATE Inherit the parameters from the previous model.
  \STATE Initialize the combination weights of the newly added operation, as well as newly added model parameters in the expanded model with network morphism or steepest descent splitting discussed in Section~\ref{Sec:3.2.2}.
  \ENDIF
  \RETURN $A_{\text{expanded}}$, $\theta_{\text{expanded}}$
\end{algorithmic}
\label{Alg:4.4}
\end{algorithm}

\section{Experimental results}\label{Sec:6.4}

The experiments will compare the existing methods with the newly proposed methods including: (a) DARTS with nodes growing and pruning at each stage (denoted as ``DARTS-1''); (b) DARTS with nodes growing and pruning at the end (denoted as ``DARTS-2''); 
For RNN models, we also perform experiments with: (d) Two-to-One back-bone architecture with nodes growing and pruning at each stage (denoted as ``new-1''); (e) Two-to-One back-bone architecture with nodes growing and pruning at the end (denoted as ``new-2''). 
The experiments will cover multi-variate time series forecasting with stacked RNN, language modeling with RNN and image classification with convolutional neural networks. The growing mechanism applied at this stage is pure node growing with network morphism based on the discussion in Section~\ref{Sec:3.2.1}, while the experiments with growing of operations will be done later. One cloud Tesla V100 and 2 Intel Xeon 8-Core
CPU are implemented for this experiment. Due to the limitation of computational power, the current experiments are based on small datasets and pure models without the tricks implemented in state-of-the-art models. Thus the results are indicative rather than conclusive.

\subsection{Experiment on multi-variate time series forecasting}\label{Sec:6.4.1}
\begin{figure}[th] 
\begin{center}
 \includegraphics[width=1.0\linewidth]{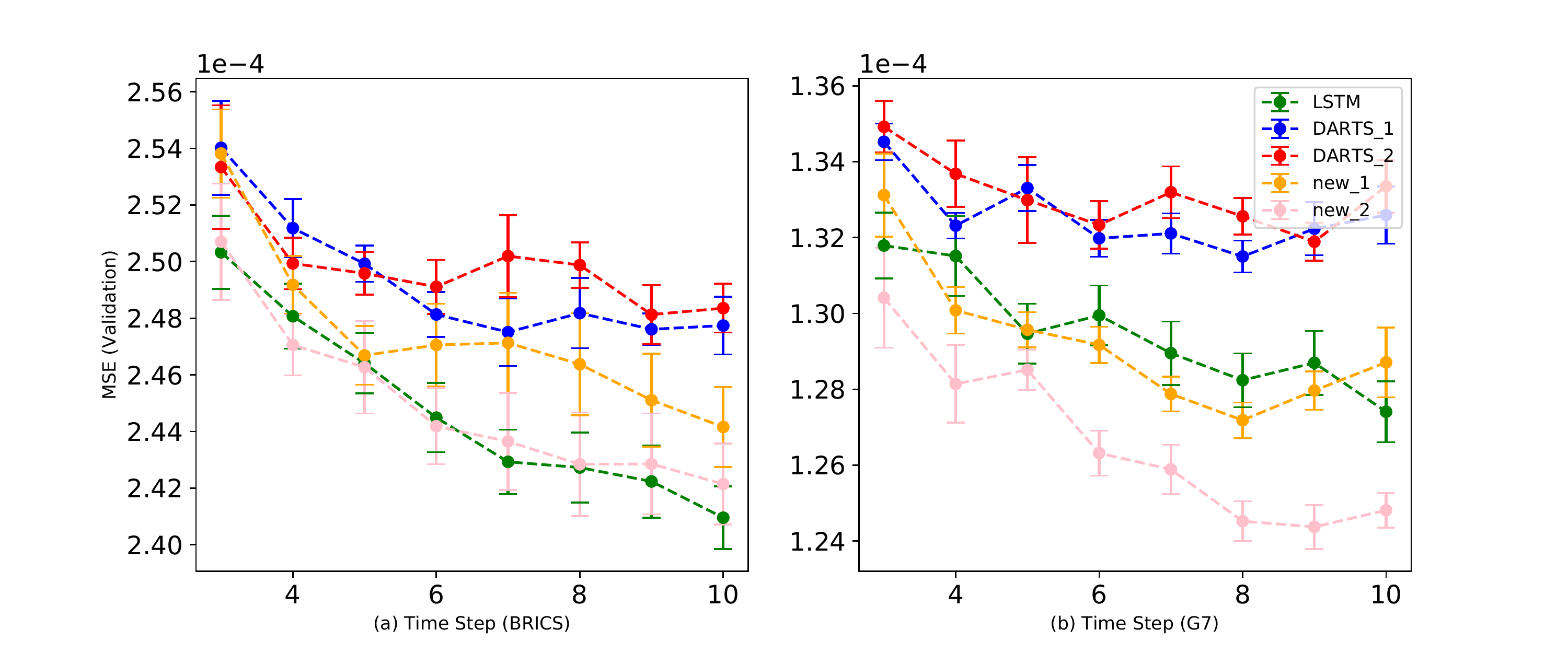}
 \caption{Comparison of the learning curves of different architectures on multi-variate time series forecasting tasks.} \label{Fig:5.3}
\end{center} 
\end{figure}

The first is a toy experiment for multi-variate time series forecasting on financial time series data. The datasets being studied are the stock indices returns of G7 countries and BRICS countries in the last ten years, while the task is one-step-ahead prediction for the return vectors of all indices involved in each case. The window size is selected to be 10 time steps, the mini-batch size is set to be 50. We employ an Adam optimizer, and the learning rate is 0.001 as is recommended in similar tasks. The updating rule of architecture parameters is based on Algorithm~\ref{Alg:4.1} with Adam optimizer. The corresponding updating rate for architecture parameter is 0.0003. We do not introduce regularization on both model parameters and architecture parameters so as to compare different models in the same condition. We compare the learning curves of RNN with different cell architectures including LSTM, GRU, Two-to-One backbone architecture, the cell architecture obtained with online growing under the mode of ``prune at each stage'', and the cell architecture obtained with the mode of ``prune at the end''. The comparison of learning curves is given by Figure~\ref{Fig:5.3}, where each curve is averaged by 20 trials with 20 selected models along with error bars for the sample standard errors.
It is shown in Figure~\ref{Fig:5.3} that with one hidden layer recurrent neural networks, LSTM works well in terms of validation MSE on both BRICS and G7 time series forecasting. Generally it outperforms both of the final pruned models from DARTS. Meanwhile, we can notice that, for both ``prune at each stage" mechanism ``prune at final" mechanism, the final models of the Two-to-One backbone significantly outperform the final model from DARTS backbone in terms of almost the whole validation learning curves. Also, with ``prune at final" mechanism, the final model from the Two-to-One backbone significantly outperforms LSTM on G7 dataset, and there is no significant difference of between it and LSTM on BRICS dataset. This indicates that the Two-to-One backbone has an advantage in searching for RNN cells in multi-variate time series forecasting, while DARTS backbone may not be easy to achieve the same level of performance. One explanation could be the utilization of element-wise multiplication operation in the proposed backbone, which is also implemented in LSTM and GRU cells. Here more theoretical work is still needed to be done. 

\subsection{Experiment on PTB and wiki-text-2 with recurrent neural networks}\label{Sec:6.4.2}
The second experiment is language modeling on Penn Tree bank and Wiki-text 2 datasets. Due to the limitation of computational power, each time we use the first 4,000 training examples in the original data set as the training set, the next 500 examples as the validation set and then the next 500 examples as the test set. We use a two-layer recurrent neural network structure with one input layer and one hidden layer with 128 unit, then we replace the cell with different backbones or final models to compare their performances. 
\newline\newline
For each dataset, we conduct 5 independent trials of searching starts from different random initialization of the backbone architectures. We show the average learning curves of the final selected models from 5 different trials on both training set and validation sets, as well as the error bars of standard error. The optimization process is the same with the last experiment, where the learning rates for model parameters and architecture parameters are still 1e-3 and 3e-4, respectively. The number of learning epochs is set to be 10. In the first simulation, we start from 2 nodes in both two NAS backbones and introduce growing mechanism after early stopping with a patience of 2 epochs. 
\begin{figure}[th] 
\begin{center}
 \includegraphics[width=1.0\linewidth]{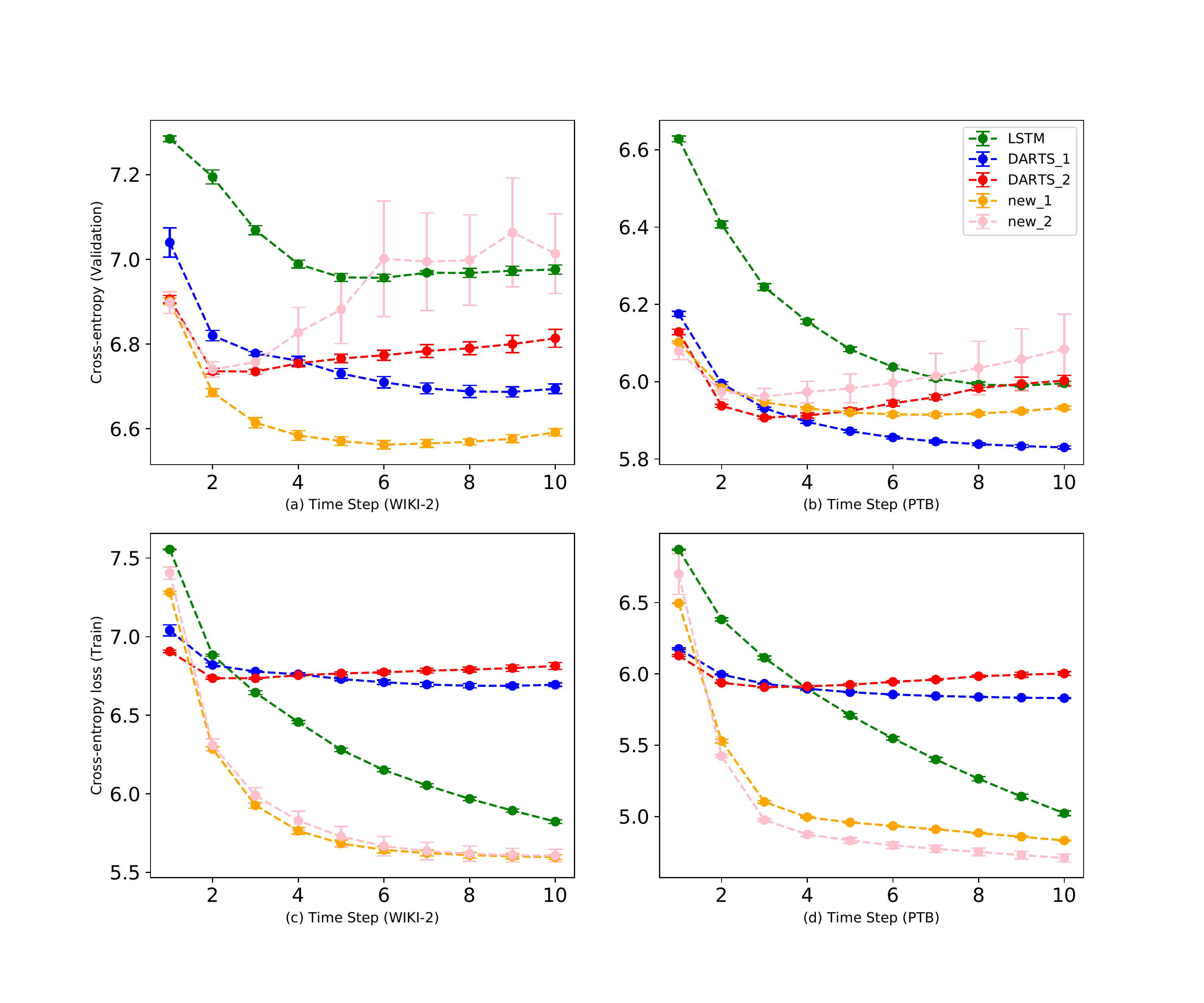}
 \caption{Comparison of the learning curves of different architectures on language modeling tasks.} \label{Fig:5.4}
\end{center} 
\end{figure}
From the learning curves on validation sets given by Figure~\ref{Fig:5.4}(a) and Figure~\ref{Fig:5.4}(b), we can see that in both cases, traditional LSTM does not perform the best even after convergence compared with the final models found by DARTS and the Two-to-One backbone. Also, for the two NAS backbones, the ``prune at each stage'' strategy gives better final models when comparing with ``prune at final'' strategy. On the other hand, by looking at the learning curves on training set given by Figure~\ref{Fig:5.4}(c) and Figure~\ref{Fig:5.4}(d), we notice that the training loss of LSTM is still going done even after 10 steps. As the decreasing of the training loss in later epochs does not bring a decreasing for the generalization performance on validation set, we can consider this decrease as the result of over-fitting. For the two NAS backbones, the training loss of the Two-to-One is generally smaller than that of DARTS.
\begin{figure}[th] 
\begin{center}
 \includegraphics[width=1.0\linewidth]{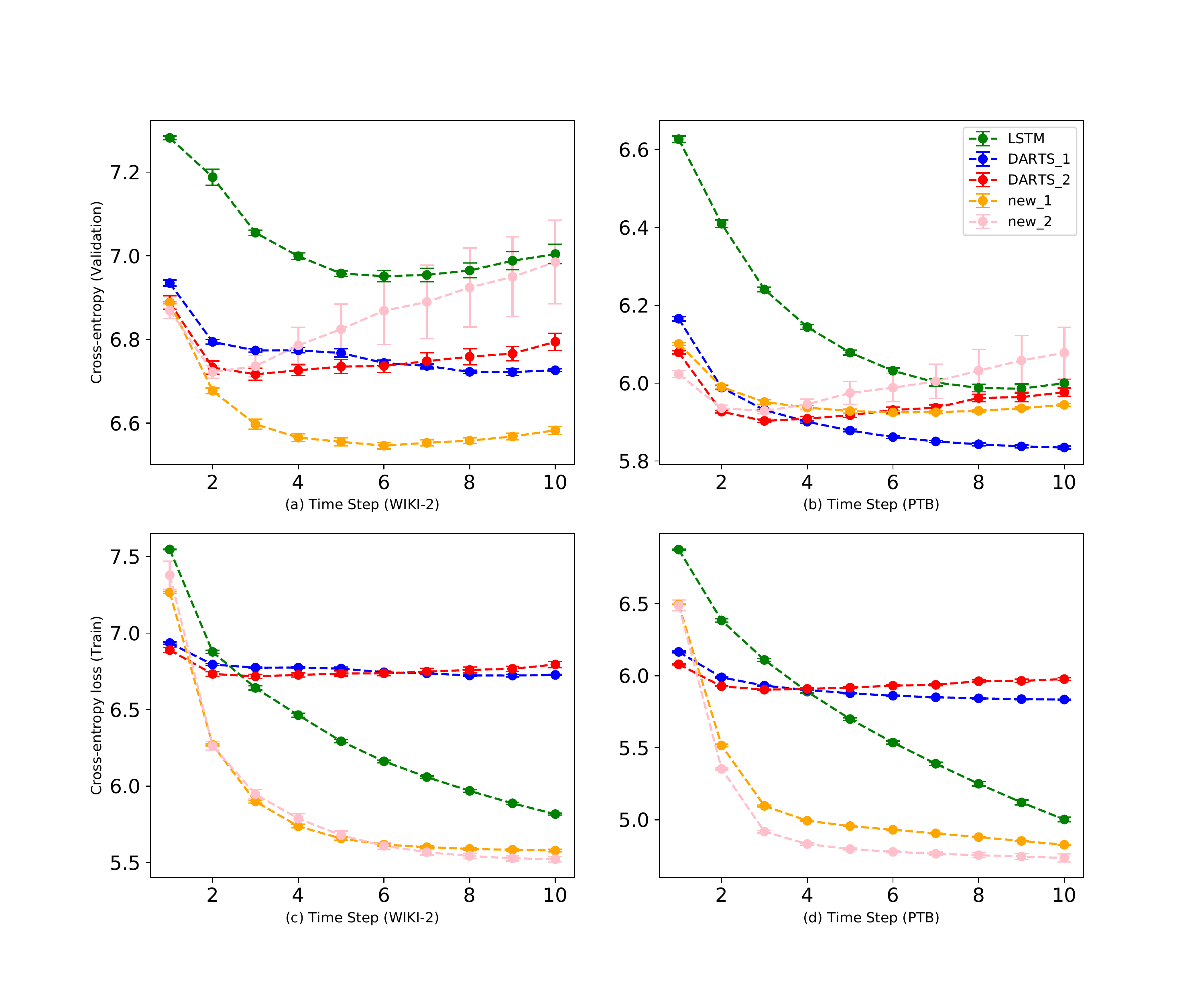}
 \caption{Comparison of the learning curves of different architectures on language modeling tasks.} \label{Fig:5.5}
\end{center} 
\end{figure}
\newline\newline
In the second simulation shown in Figure~\ref{Fig:5.5}, we start from 3 nodes in both two NAS backbones and introduce growing mechanism after early stopping with a patience of 3 epochs. Actually the comparison of validation and test performances between final models are the similar with Figure~\ref{Fig:5.4}. Especially the two backbone architectures give better validation performances when applied with ``prune at each stage'' strategy, compared with LSTM and other settings. Also, the test performance of Two-to-One backbone outperform that of DARTS' for both two pruning mechanisms.
\begin{figure}[th] 
\begin{center}
 \includegraphics[width=1.0\linewidth]{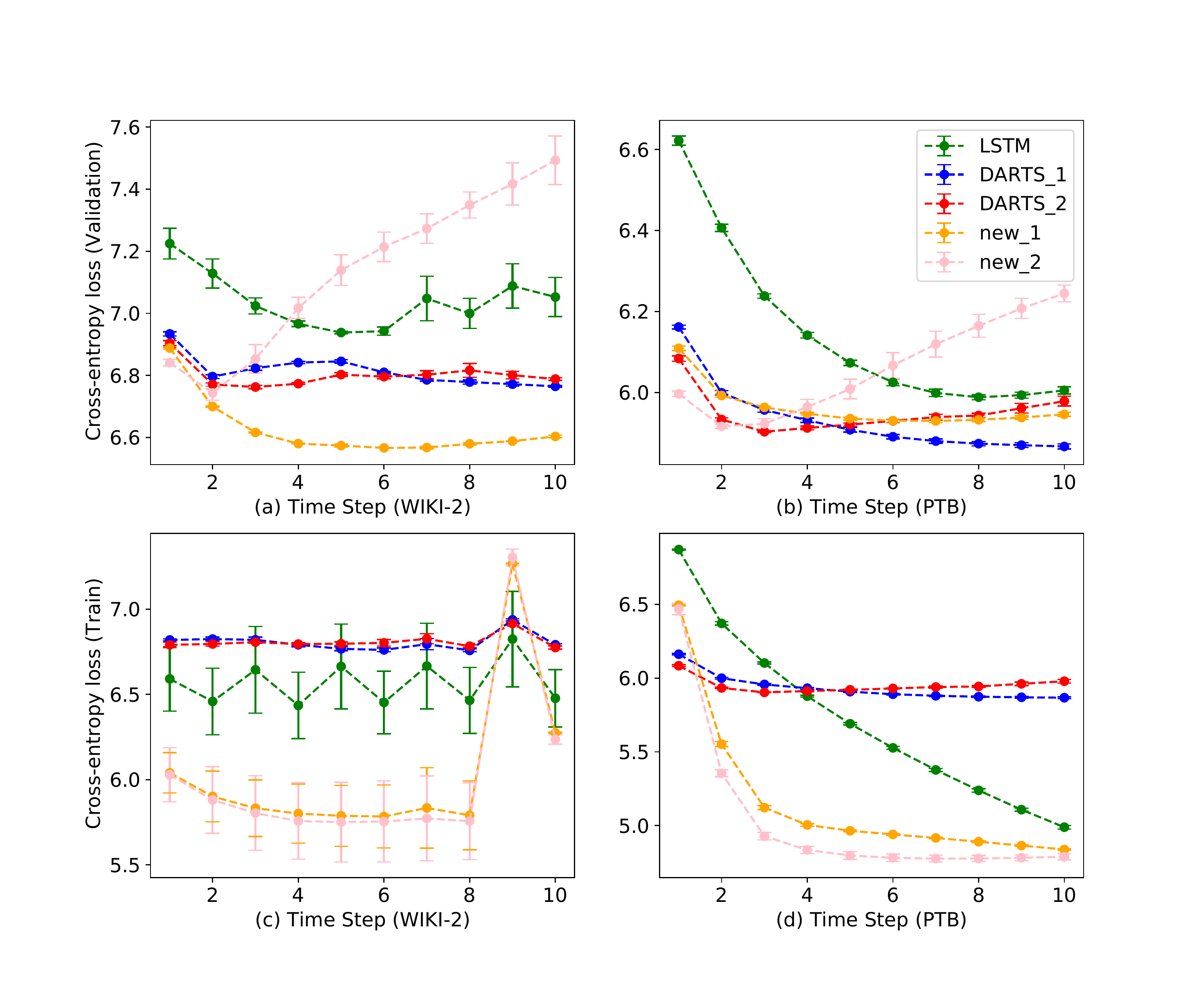}
 \caption{Comparison of the learning curves of different architectures on language modeling tasks.} \label{Fig:5.6}
\end{center} 
\end{figure}
\newline\newline
In the simulation with the results shown by Figure~\ref{Fig:5.6}, we use a larger starting number of nodes $n=4$ for two NAS backbones. We still compare them with the same LSTM setting. We find that for most of the time, the proposed ``prune at each stage'' mechanism works better than traditional ``prune at the end'' on validation sets for both two benchmark datasets. On the other hand, the proposed two-input backbone in general outperforms DARTS architecture on training sets for both these two benchmarks datasets.
\newline\newline
Next we compare the intermediate validation performance with respect to real computing times for different NAS methods and different starting number of nodes with growing mechanism. In each simulation, we run each methods for 5 times, and record the validation performance at each epoch with corresponding computing time. 
\begin{figure}[th] 
\begin{center}
 \includegraphics[width=1.0\linewidth]{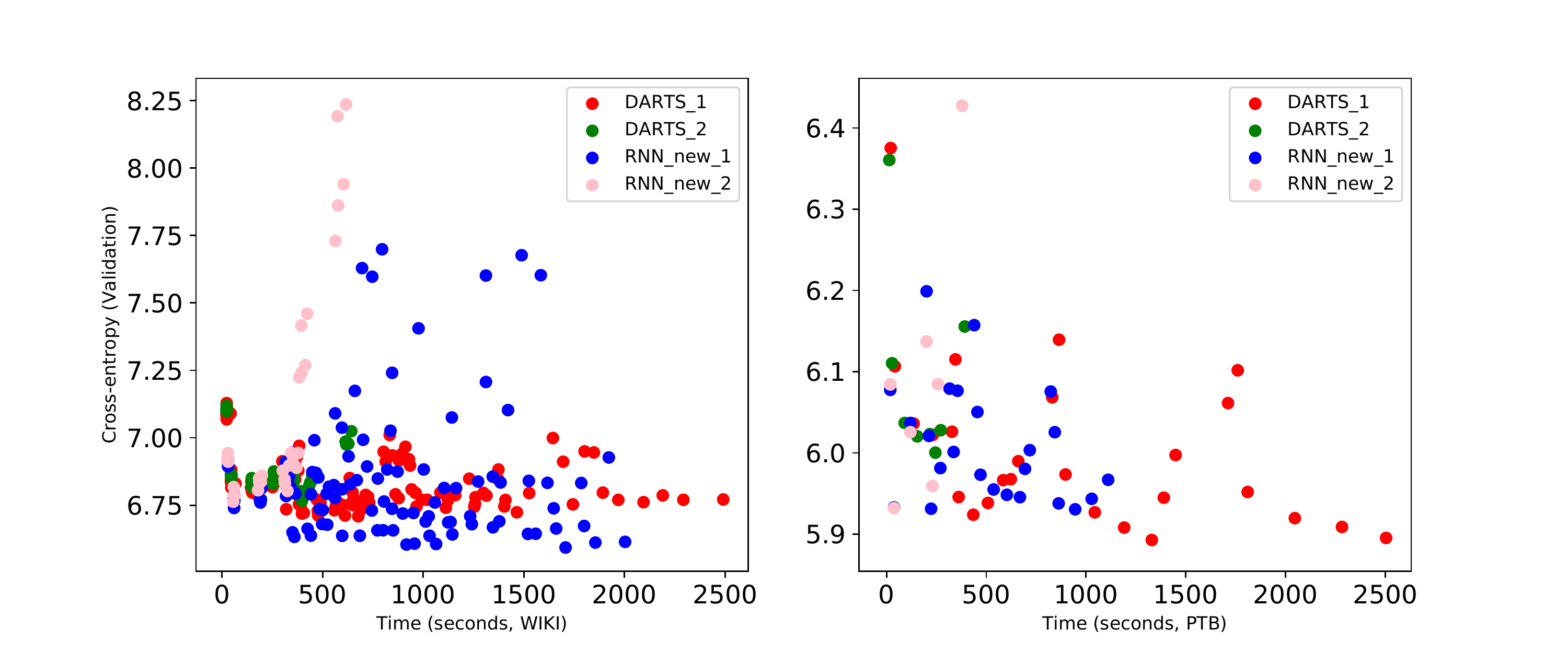}
 \caption{Comparison of different architectures searching strategy on language modeling tasks in real training times.} \label{Fig:5.7}
\end{center} 
\end{figure}
\newline\newline
Since we combined multiple stage of training and multiple trials in a single figure, the data points does not appear with clear trends. It is shown in Figure~\ref{Fig:5.7} that by comparing the four NAS methods. In general, the Two-to-One backbone works the most efficiently on WIKI-text-2 with ``prune at each stage" mechanism, followed by the DARTS backbone with ``prune at each stage" mechanism. The whole training process for ``prune at each stage'' mechanism is longer than ``prune at final'' mechanism as we need to retrain the model after each time of pruning. However, the observation points of the ``prune at each stage" mechanism can achieve lower validation performances or at least no worse than the ``prune at final'' mechanism even at early time points, while its optimal performances are almost always better.
\begin{figure}[th] 
\begin{center}
 \includegraphics[width=1.0\linewidth]{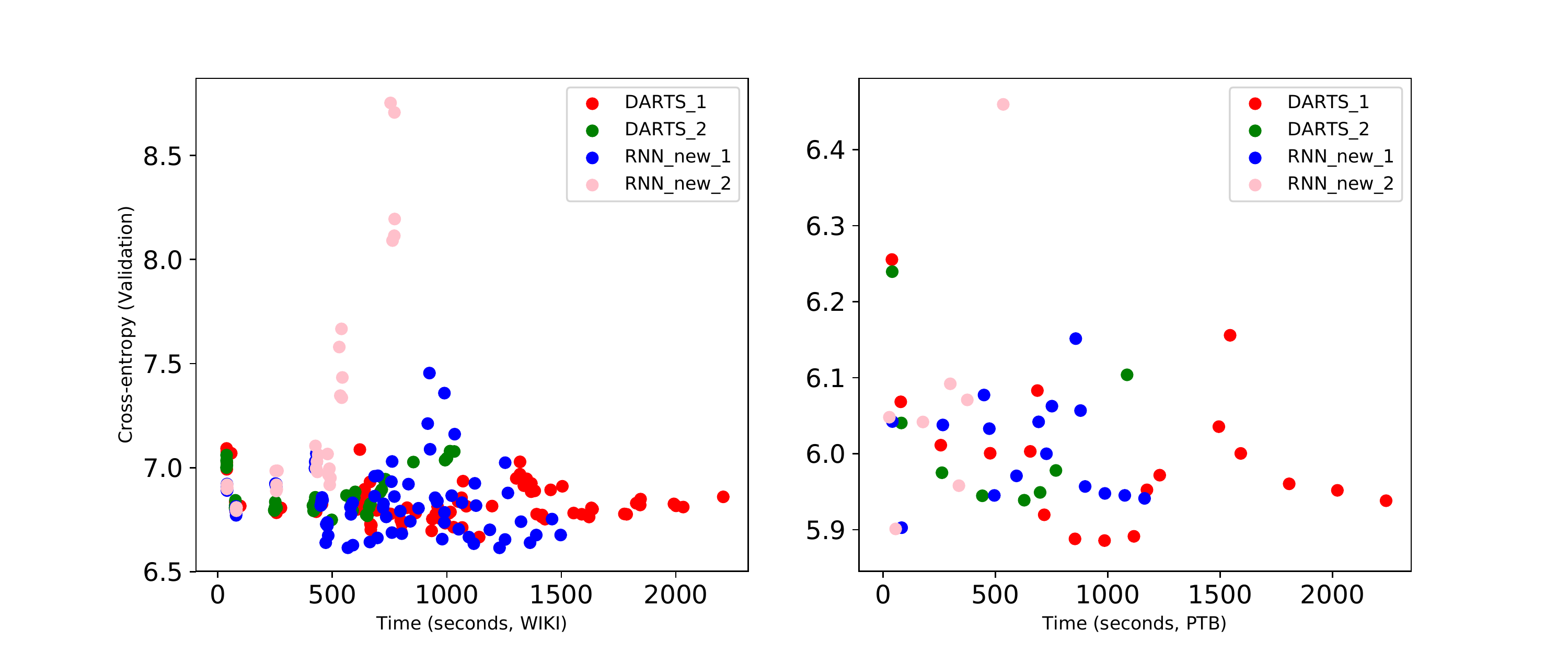}
 \caption{Comparison of different architectures searching strategy on language modeling tasks in real training times.} \label{Fig:5.8}
\end{center} 
\end{figure}
In the second simulation for the training process with real computing times shown in Figure ~\ref{Fig:5.8}, we start from 3 nodes in both two NAS backbones and introduce growing mechanism after early stopping with a patience of 3 epochs. 
\begin{figure}[th] 
\begin{center}
 \includegraphics[width=1.0\linewidth]{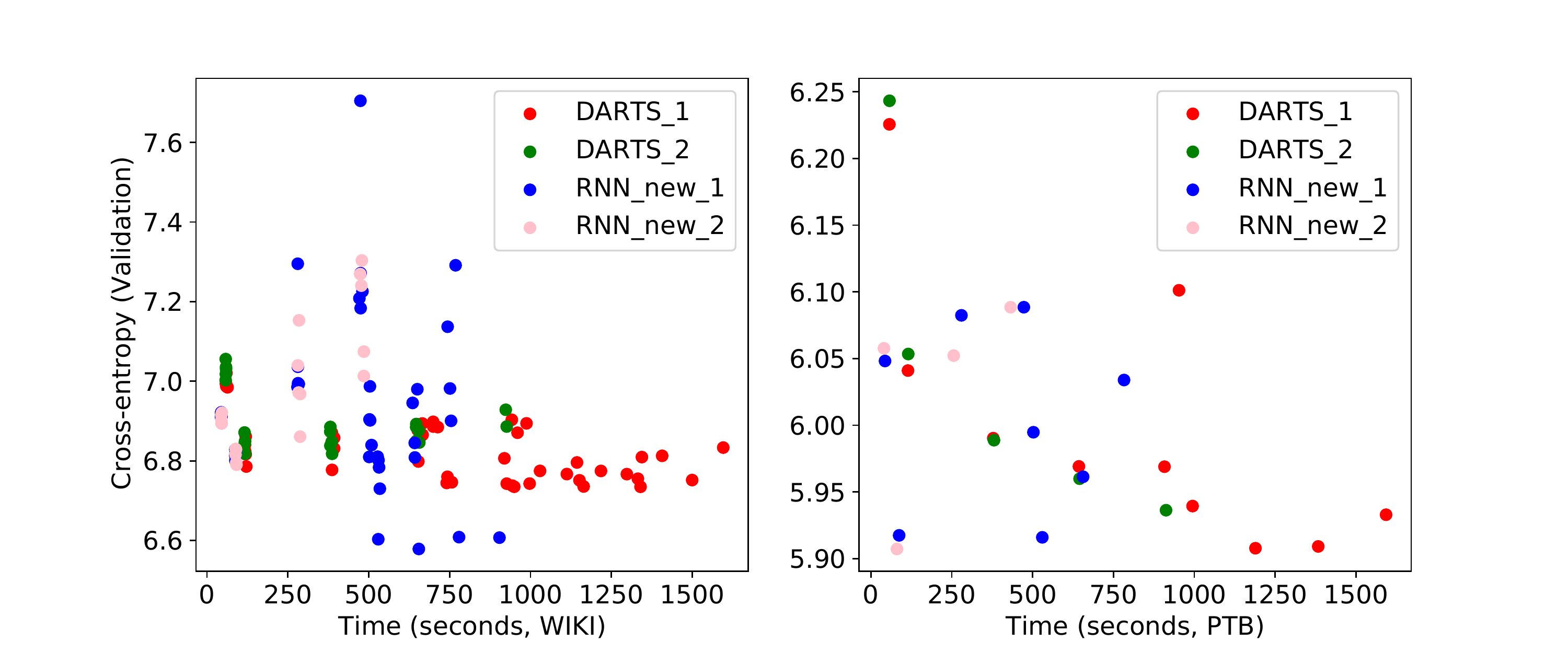}
 \caption{Comparison of different architectures searching strategy on language modeling tasks in real training times.} \label{Fig:5.9}
\end{center} 
\end{figure}

In the third simulation for the training process with real computing times shown in Figure~\ref{Fig:5.9}, we start from 4 nodes in both NAS backbones and introduce growing mechanism after early stopping with a patience of 3 epochs. 
\newline\newline
We notice that for all the three cases and both WIKI-text-2 and PTB datasets, the proposed ``prune at each stage'' for both DARTS and Two-to-One backbone architecture achieve better performance than the traditional ``prune at the end'' strategy. Also, with the different starting numbers of nodes, the proposed growing and prunning mechanisms for DARTS or Two-to-One have the potential to find architectures that outperform classical LSTM for RNN networks in the two learning tasks.


\section{Discussion}\label{Sec:6.5}
\subsection{Number of parameters}\label{Sec:6.5.1}
For typical RNN models, the number of parameters depends on the type of hidden cells used, in general it is $N = g\sum_{i=1}^{L-1} (n_{i+1}(n_{i+1}+n_i) +n_{i+1})$, where $g$ is the number of weight matrices in each cell, $n_1$ is the number of input elements, and $n_i (i>1)$ is the number of hidden units in different hidden layers. For the basic RNN, $g=1$, for GRU, $g=3$, and for LSTM, $g=4$.
\newline\newline
For DARTS backbone, we have 5 operations for each node. Among them, linear mapping followed by tanh, sigmoid or relu will introduce a weight matrix with input dimension $n_x + n_h$ and output dimension $n_h$. Assume that we have $m$ nodes, the first node will have $3\times(n_x + n_h)n_h$ parameters, while each of the rest $m-1$ nodes has $\sum_{k=2}^m 3(k-1)\times n_h^2 $ parameters. Also, for each nodes we have 5 extra parameters for combination weights. Therefore, the number of parameters in a DARTS backbone cell will be
\begin{equation}
    N_{p, \text{darts cell}}  = 3(n_x + n_h)n_h + \sum_{k=2}^m 3(k-1) n_h^2 + 5m
    \label{Eq:6.DARTScell number of parameters}
\end{equation}
where we only consider the case of one hidden layer. When the model have multiple layers, the total number of parameters is given by:
\begin{equation}
    N_{p, \text{darts total}}  = \sum_{j=1}^{L-1}(3(n_j + n_{j+1})n_{j+1} + \sum_{k=2}^m 3(k-1) n_{j+1}^2 + 5m)
\end{equation}
where $L$ is the total number of layers, and $n_j$ is the number of states in $j$th layer. For the proposed backbone, we also have 5 but different two-input operations for each node. Among them, linear combination followed by tanh, sigmoid or relu will introduce two weight matrices with size $n_x n_h$ and $n_h^2$ as well as a bias with size $n_h$ for each node, while sum and element-wise product will introduce two weight matrices with size $n_x n_h$ for each node. Therefore, for a backbone with $m$ nodes, the total number of parameters is:
\begin{equation}
    N_{p, \text{new cell}} = m [3(n_x n_h + n_h^2 + n_h) + 2 n_x n_h] + 5m
    \label{Eq:6.new cell number of parameters}
\end{equation}
The total number of parameters in the architecture is given by
\begin{equation}
    N_{p, \text{new total}} = \sum_{j=1}^{L-1} (m [3(n_j n_{j+1} + n_{j+1}^2 + n_{j+1}) + 2 n_j n_{j+1}] + 5m)
\end{equation}
Now take the settings in G7 indices prediction in Section~\ref{Sec:6.4.1} for example, where we implemented a single layer RNN with input size and hidden size both to be 7. The summary and examples of number of parameters in different architectures or backbone architectures are given in Table~\ref{tab:5.number of parameters}.
\begin{table}[htbp]
  \centering
  \caption{Number of parameters in different architectures or backbone architectures}
    \begin{tabular}{cccrr}
    \toprule
    Model & Nodes & Number of parameters & \multicolumn{1}{c}{G7} & \multicolumn{1}{c}{BRICS} \\
    \midrule
    LSTM  &       &   $ 4(n_h(n_h+n_x) +n_h)$    & 392   & 200 \\
    \midrule
    GRU   &       &   $ 3(n_h(n_h+n_x) +n_h)$    & 294   & 150 \\
    \midrule
    \multirow{5}[2]{*}{DARTS} & 2     & \multirow{6}[2]{*}{} & 451   & 235 \\
          & 3     &  & 750   & 390 \\
          & 4     &   $3(n_x + n_h)n_h + \sum_{k=2}^m 3(k-1) n_h^2 + 5m$   & 1196  & 620 \\
          & 5     &       & 1789  & 925 \\
          & 6     &       & 2529  & 1305 \\
          & 7     &       & 3416  & 1760 \\
    \midrule
    \multirow{5}[2]{*}{RNN\_new} & 2     & \multirow{6}[2]{*}{} & 836  & 440 \\
          & 3     &   & 1254  & 660 \\
          & 4     &  $m [3(n_x n_h + n_h^2 + n_h) + 2 n_x n_h] + 5m$     & 1672  & 880 \\
          & 5     &       & 2090  & 1100 \\
          & 6     &       & 2508  & 1320 \\
          & 7     &       & 2926  & 1540 \\
    \bottomrule
    \end{tabular}%
  \label{tab:5.number of parameters}%
\end{table}%

As we can see, by comparing the number of parameters for DARTS and Two-to-One backbone architecture, we notice that when the number of nodes is small, DARTS backbone has less number of parameters as it has two candidate operation without any parameter. However, since DARTS backbone is a fully connected graph, there is one term in Eq.~\eqref{Eq:6.DARTScell number of parameters} that is in second-order proportional to the number of nodes, while in Eq.~\eqref{Eq:6.new cell number of parameters} only one-order terms for node number are included. This indicates that when the number of node becomes larger, the total number of parameters in DARTS will become increasingly larger than that of Two-to-One backbone. On the other hand, as over-parameterized graphs, backbone architectures are usually larger than conventional cell architectures, while the number of model parameters after pruning is determined by which operation is kept in each node.

\subsection{Computational complexity}\label{Sec:6.5.2}
As we know, typical LSTM or GRU models are local in space and time \citep{hochreiter1997long}. Its computational complexity per time step and weight is $O(1)$, while the overall complexity of an LSTM per time step is equal to $O(w)$, where $w$ is the number of weights \citep{tsironi2017analysis}. The cases of DARTS and Two-to-One backbone are similar. Now we consider the computational complexity for growing of cell structure and training a fixed but larger backbone.
\newline\newline
First, assume that we start training from a $m_0$-node backbone cell structure which will be saturated in $e_0$ epochs. Also we assume each later stage after adding new node requires $e_i\,(i>0)$ epochs to early stop. For DARTS backbone the computational complexity is given by:
\begin{equation}
\begin{split}
    C_{DARTS, grow} &= O(\sum^{M-m_0}_{i=0} N_i e_i) = O(e_0 (3(n_x + n_h)n_h + \sum_{k=2}^{m_0} 3(k-1) n_h^2 + 5m_0)\\
    &+ \sum_{i=1}^{M-m_0} e_i (3(n_x + n_h)n_h + \sum_{k=2}^{m_i} 3(k-1) n_h^2 + 5m_i))
    \end{split}
\end{equation}
For the Two-to-One backbone:
\begin{equation}
\begin{split}
    C_{new, grow} &= O(\sum_{i=0}^{M-m_0} N_i e_i) = O(e_0 (m_0 [3(n_x n_h + n_h^2 + n_h) + 2 n_x n_h] + 5m_0)\\
    &+ \sum_{i=1}^{M-m_0} e_i (m_i [3(n_x n_h + n_h^2 + n_h) + 2 n_x n_h] + 5m_i))
    \end{split}
\end{equation}
On the other hand, for fixed backbones with $M$ nodes, the corresponding training computational complexity is given by:
\begin{equation}
\begin{split}
    C_{DARTS, \text{fixed}}&=O(3(n_x + n_h)n_h + \sum_{k=2}^M 3(k-1) n_h^2 + 5M)\\
    C_{new, \text{fixed}}&=O(M [3(n_x n_h + n_h^2 + n_h) + 2 n_x n_h] + 5M)
    \end{split}
\end{equation}
Generally, if $C_{DARTS, grow} < C_{DARTS, fixed}$ or $C_{new, grow} < C_{new, fixed}$, introducing cell backbone growing mechanism will reduce the searching time. In most of the cases, we do not know the optimal values of $M$. Therefore, even though the corresponding computational complexity of training with growing mechanism could be larger, it can be helpful in avoiding the search of $M$ and thus reducing the whole searching time. 
\section{Conclusion}\label{Sec:6.6}
In this study, we proposed a method for online architecture adaptation by combining gradient-based neural architecture search and dynamic network transformation. Two transformation mechanisms are introduced: The first is growing new nodes based on early stopping criterion and network morphorism, while the second is dynamically growing and pruning operations during the training process with network morphorism. We also introduced a novel two-input backbone cell architecture for recurrent neural networks that keeps identity mappings for both input and hidden states. Further, we provide a novel aspect of combining two techniques for one shot architecture search including differentiable architecture search and dynamic network with growing and pruning mechanism. Initial experiment on time series forecasting and language modeling indicates that the Two-to-One backbone has potential to achieve similar or even stronger performance than the traditional LSTM model and the existing DARTS backbone as recurrent cells. Also, further investigation suggests that, by introducing growing mechanism that training backbone from smaller to larger ones with network morphism, the total searching time of satisfactory cell structure can be largely reduced in certain cases. In addition, we implemented analytical derivation to compare the computational complexities of DARTS backbone and the Two-to-One backbone. We found that as the number of nodes becomes larger, the computational time cost of our proposed backbone increases much slower than DARTS backbone, while there exists a break-even point around five or six nodes in the examples of muti-variate time series forecasting on G7 and BRICS indices.

\newpage
\bibliography{research}

\end{document}